# OCR ACCURACY IMPROVEMENT ON DOCUMENT IMAGES THROUGH A NOVEL PRE-PROCESSING APPROACH

A. El Harraj[1] and N. Raissouni[2]

[1,2]RSAID Laboratory: "Remote sensing/Signal-image Processing & Applied mathematics/Informatics/ Decision making". The National School for Applied Sciences of Tetuan. Univeristy of Abdelmalek Essaadi. BP. 2222. M'Hannech II. 93030. Tetuan. Morocco.

## ABSTRACT

*Digital camera and mobile document image acquisition are new trends arising in the world of Optical Character Recognition and text detection. In some cases, such process integrates many distortions and produces poorly scanned text or text-photo images and natural images, leading to an unreliable OCR digitization. In this paper, we present a novel nonparametric and unsupervised method to compensate for undesirable document image distortions aiming to optimally improve OCR accuracy. Our approach relies on a very efficient stack of document image enhancing techniques to recover deformation of the entire document image. First, we propose a local brightness and contrast adjustment method to effectively handle lighting variations and the irregular distribution of image illumination. Second, we use an optimized greyscale conversion algorithm to transform our document image to greyscale level. Third, we sharpen the useful information in the resulting greyscale image using Un-sharp Masking method. Finally, an optimal global binarization approach is used to prepare the final document image to OCR recognition. The proposed approach can significantly improve text detection rate and optical character recognition accuracy. To demonstrate the efficiency of our approach, an exhaustive experimentation on a standard dataset is presented.*

## 1. INTRODUCTION

Text based information systems have become increasingly important in almost all fields. In many situations (such as physical newspapers or old printed books), the source of the input text is not from an editable documents, but instead documents in their original paper form. In some cases imaging systems can be used to store and retrieve these documents through manually assigned key words, but full text access can be more effective as it will enable an automated process for storing, indexing and information retrieving with full access to all content key words. In order to get full-text content from paper documents Optical Character Recognition (OCR) is used. For scanned documents, OCR techniques can recognize words with a high level of accuracy and so can be used to extract and/or index information for further information retrieval. However, automatically extracting text from document images produced by a digital camera or a mobile camera is a very challenging task. Enabling text extraction from these documents with a high accuracy will provide the enabling technology for a number of applications such as improved indexing, online searchable documents, indoor/outdoor scene understanding [12], and text detection in natural images [13]. OCR technology offers better search and retrieval functionality than has been possible before [14]. As opposed to the classical scanned documents, text detection in document images produced by digital cameras and mobile cameras is a very challenging task, far from being completely solved. Complex backgrounds, uneven illumination, and presence of almost unwanted text fonts, sizes, and orientations pose great difficulties even to state-of-the-art text detection methods.

In this paper we focus on developing an approach that solves the problem of extracting text from document images. Our approach has several advantages. First, we propose to use a new model for illumination adjustment based on Contrast Limited Adaptive Histogram Equalization (CLAHE) to improve the contrast of the overall objects presents in the processed document image. Second, we use Luminance algorithm to optimize grayscale conversion for text extraction. Third, to enhance text details and edges we using Un-sharp masking filter. As a final enhancement step we use Otsu Binarization algorithm as a powerful method for Cleaning and whitening document background. The proposed method is evaluated using a standard datasets available on [42].

The article is organized as follows. The proposed algorithm is presented in detail in Section 3, successively, we present the adopted models for Illumination adjustment, contrast conversion, text details sharpening and finally document background cleaning and whitening. Section 4 introduces the setup environment used to evaluate our approach and the used datasets, and presents the evaluation results for the proposed method. Finally, the article is concluded in Section 5.

## 2. RELATED WORK

Several approaches have been proposed to enhance OCR accuracy and text detection. Some approaches tried to correct OCR errors after detecting them. In [43] Kukich proposed to use a dictionary or n-gram based approaches to detect OCR errors and replace them with the most likely word in the dictionary using statistical measures. These approaches can reduce the overall OCR error rates for the frequent words of the language, but it is likely to corrupt correctly recognized words which are not in the dictionary for instance names and places. As an alternative Tong et al [44] proposed to use the context of the text itself to correct misrecognized words. Nevertheless, the success of these approaches depends on the language models and trained dictionaries and can be useless if used on different corpora with different vocabularies. A more recent alternative is to combine multiple OCR outputs to locate and fix OCR errors automatically without using language specific information [45].

Other approaches are based on edge detection, binarization, connected-component based and texture-based methods [15] [16]. In [15], the authors demonstrate that best results were achieved using edge-based text detection compared to mathematical morphology and color-based character extraction. Other researchers tried to extract text from road signs [17], license plates [18], library books [19], web images [20], video frames [21] [24], natural image scenes [25][26][27] . Moalla et al. [22] developed a method to classify medieval manuscripts by different scripts in order to assist paleographers. Ben Jlaiel et al. [23] suggested a strategy to discriminate Arabic and Latin modern scripts that can be applied also to ancient scripts. J. Edwards et al. [28], on the other hand, developed a method based on a generalized Hidden Markov Model to improve accuracy on Latin manuscripts up to 75%. Other researchers proposed to improve OCR accuracy through post-processing techniques on the output of a single or multiple OCR engines [29][30][31][32][45]. More recently, some researchers tried to get good OCR accuracy from document images [33] [34] by removing degradation problems, noise and improving the quality of the documents [35].

In this particular work, we focus on improving OCR accuracy by pre-processing the input documents images. Contrary to the method proposed in [35], we propose to use a slightly more enhanced non parametric approach by improving the quality of the input document, thus using a tested combination of pre-processing techniques. This gives us the possibility to improve OCR accuracy independently to any trained dictionary, language specific information or language model.

## 3. THE PROPOSED METHOD

In this section, we exhaustively detail the key stages in our approach, explicitly, Illumination adjustment, grayscale conversion, Un-sharp masking and optimized adaptive thresholding for binarization. We note that our method can be used with any kind of document images.

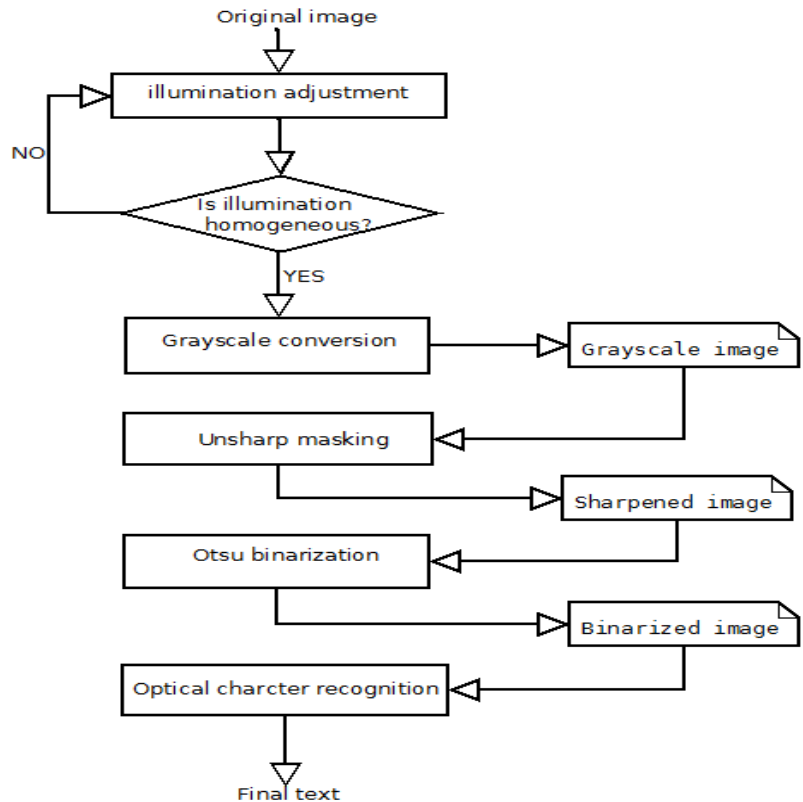

Figure 1: Diagram of different steps involved in the proposed approach.

## 3.1. Illumination adjustment

The first step in our preprocessing stack is to improve the contrast of the overall objects presents in the processed image. For that raison, we will use CLAHE [11] as a powerful contrast enhancement.

Contrast enhancement methods are not intended to increase or supplement the intrinsic structural information in an image but rather to improve the image contrast and hypothetically to enhance particular characteristics [11]. The input images are 8-bit grayscale images. We can process these images directly. But there is a slight problem with that. Black-to-White transition is taken as Positive slope (it has a positive value) while White-to-Black transition is taken as a Negative slope (It has negative value). So when we convert data, all negative slopes are made zero. And then we miss some edges. To bypass this problem, we convert data type to some higher forms like 16-bit, 64-bit etc (we use 16-bit grayscale images), process it and then convert back to original 8-bit.

### 3.1.1. Brightness Equalization

As a simple solution, we can convert our input image to HSV color space and search for the brightness value using the V channel. The HSV color space has three components: hue, saturation and value.

HSV hue is a number in the interval [0, 360]. The hue of any neutral color--white, gray, or black--is set at 0°.

HSV Saturation measures how close a color is to the grayscale. S ranges from 0 to 1. White, gray, and black all have a saturation level of 0. Brighter, purer colors have saturation near 1.

HSV Value is the highest value among the three R, G, and B numbers. This number is divided by 255 to scale it between 0 and 1. In terms of perception, HSV Value represents how light, bright, or intense a color is. Value does not distinguish white and pure colors, all of which have V = 1. 'Value' is sometimes substituted with 'brightness' and then it is known as HSB [36] [37][38]. Unfortunately this approach is not very efficient for brithness estimation. A more efficient solution would be the use of local bi-histogram equalization (LBHE) [40]. This approach was successfully tested for the motion and brightness estimation in MPEG system [40]. When we tested this approach, we concluded that this approach is not suited for document images. But instead, we use the decomposition of HSV space and apply Contrast Limited Histogram Equalization (CLAHE) [11] to the Value (V) for brightness equalization.

CLAHE is used as a very efficient adaptive histogram equalization that's has demonstrated to be successful for enhancement of low-contrast images such as portal films [11]. CLAHE is based on Adaptive Histogram Equalization (AHE) [11]. AHE calculate the histogram for the contextual region of a pixel. Then transform the resulting pixel's intensity to a value within the display range proportional to the pixel intensity's rank in the local intensity histogram but this process can over amplify the noise in the initial image. CLAHE overcome this issue by refining AHE by imposing a user specified maximum, ie, Clip Limit, to, the height of the local histogram, and thus on the maximum contrast enhancement factor. The enhancement is thereby reduced in very uniform areas of the image "tiles" [11]. The resulting neighboring tiles are then stitched back seamlessly using bilinear interpolation, which prevent over enhancement of noise and reduce the edge-shadowing effect of unlimited AHE (Figure 2). Thus, CLAHE can limit the noise whereas enhancing the contrast [11].

In our case we use a Uniform distribution with a clip limit equal to 4. (Figure 2) shows an example of the produced enhanced grayscale image by applying the CLAHE enhancement.
The clip limit can be obtained by: ß [11].

$$\beta = \frac{M}{N}\left(1 + \frac{\alpha}{100}(S_{max} - 1)\right) \qquad (1)$$

Where $\alpha$ is clip limit factor, M region size, and N is grayscale value. The maximum clip limit is obtained for $\alpha=100$.

The uniform CLAHE equalization is obtained by (2)

$$I = (I_{max} - I_{min}) * P(f) + I_{min} \qquad (2)$$

Where:
: computed pixel value
$I_{max}$ : Maximum pixel value
$I_{min}$ : Minimum pixel value
$P(f)$ :  Cumulative probability distribution
Let R, G and B three numbers in the interval [0, 255].

The Hue is defined as (3)

$$\begin{cases} H = \cos^{-1}[(R - \frac{1}{2}G - \frac{1}{2}B)/\sqrt{R^2 + G^2 + B^2 - RG - RB - GB}] \text{ if } G \geq B \\ 360 - \cos^{-1}[(R - \frac{1}{2}G - \frac{1}{2}B)/\sqrt{R^2 + G^2 + B^2 - RG - RB - GB}] \text{ if } B > G \end{cases} \qquad (3)$$

We define m and M as (4)

$$\begin{cases} m = \min\{R, G, B\} \\ M = \max\{R, G, B\} \end{cases} \qquad (4)$$

From (4) S and V can be defined as (5)

$$\begin{cases} V = M/255 \\ S = 1 - m/M \text{ if } M > 0 \\ S = 0 \text{ if } M = 0 \end{cases} \qquad (5)$$

First we split every channel separately and apply the CLAHE equalization to the V channel, and then we merge the color planes back into an HSV image. Finally we convert back the HSV image to RGB color space to continue the processing steps.

Let H, S, and V be the Hue, Saturation and Value in HSV color space.

From (5) we have:

$$\begin{cases} M = 255V \\ m = M(1-S) \end{cases} \quad (6)$$

We define a new variable as (7)

$$z = (M-m)[1 - |\left(\frac{H}{60}\right)\%2 - 1|] \quad (7)$$

Then R, G and B can be calculated depending on the Hue values:

$$\text{If } 0 \le H < 60 \text{ Then} \begin{cases} R = M \\ G = z + m \\ B = m \end{cases} \quad (8)$$

$$\text{If } 60 \le H < 120 \text{ Then} \begin{cases} R = z + m \\ G = M \\ B = m \end{cases} \quad (9)$$

$$\text{If } 120 \le H < 180 \text{ Then} \begin{cases} R = m \\ G = M \\ B = z + m \end{cases} \quad (10)$$

$$\text{If } 180 \le H < 240 \text{ Then} \begin{cases} R = m \\ G = z + m \\ B = M \end{cases} \quad (11)$$

$$\text{If } 240 \le H < 300 \text{ Then} \begin{cases} R = z + m \\ G = m \\ B = M \end{cases} \quad (12)$$

$$\text{If } 300 \le H < 360 \text{ Then} \begin{cases} R = M \\ G = m \\ B = z + m \end{cases} \quad (13)$$

In (figure 2) we give the resulting enhanced document image (the image is from [41]) when applying this procedure. As explained previously, we convert the RGB image to HSV color space using (3), (4) and (5) equations. Then we split the three channels separately. We apply the CLAHE equalization to the V channel then we merge back the channels to obtain the resulting HSV enhanced document image. Finally we convert back the resulting image to the RGB color space to get the final image presented in the right of figure (2).

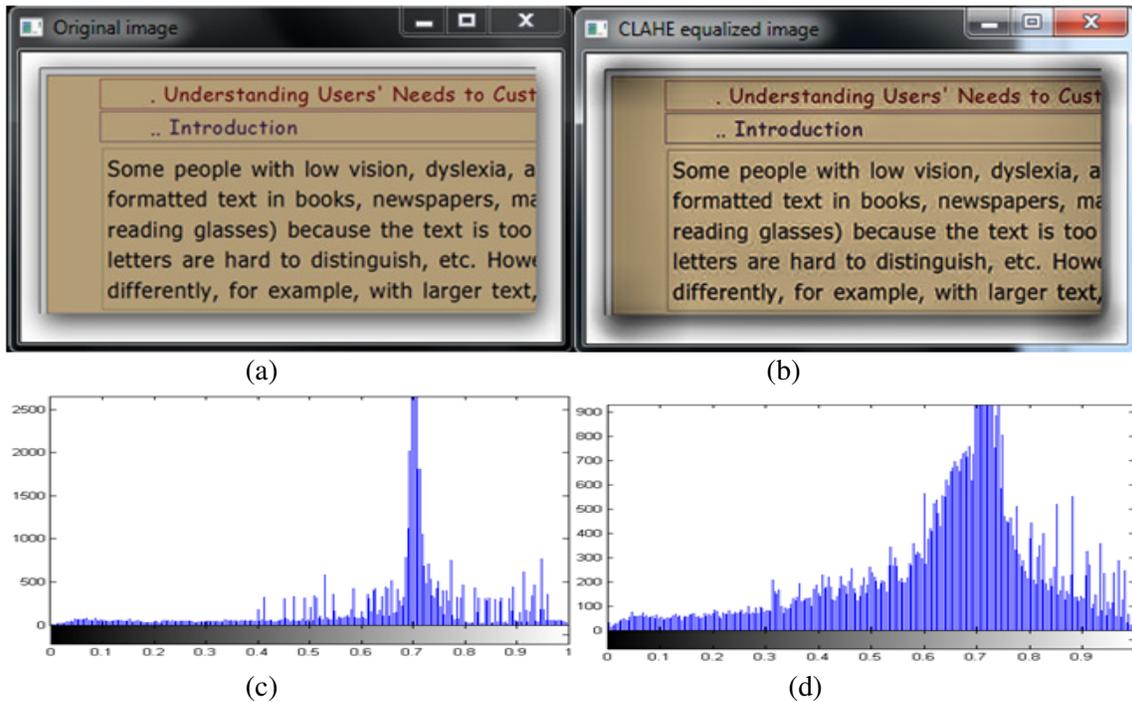

Figure 2: (a)- the original image, (b)- the enhanced image using CLAHE equalization applied on V channel for HSV Color space, (c)- Histogram of V channel for the original image converted to HSV color space, (d)- Histogram of V channel for the enhanced image converted to HSV color space

### 3.1.2. Brightness estimation

We experiment within RGB, LAB, HSV and Y'UV color space. We found that Y'UV is more suitable for our case to estimate the brightness. Y'UV is a color space encoding color images or videos taking human perception into account. Luma (Y') is a more adapted channel for brightness estimation, using the weighted average of gamma-corrected R, G, and B, based on their contribution to perceived luminance [38][39]. The relationship can found using the matrix bellow (14)

The relationship can found using the matrix bellow (14)

$$\begin{bmatrix} Y' \\ U \\ V \end{bmatrix} = \begin{bmatrix} 0.299 & 0.587 & 0.144 \\ -0.14713 & -0.28886 & 0.436 \\ 0.615 & -0.51499 & -0.10001 \end{bmatrix} \begin{bmatrix} R \\ G \\ B \end{bmatrix} \tag{14}$$

Where

$$Y' = 0.299 * R + 0.587 * G + 0.144 * B \tag{15}$$

$Y'$ is the Luma channel.

R, G and B represent the red, green and blue channels respectively.

To estimate the brightness of our image we calculate the mean average value of the Luma channel. We use this value to estimate the gain and bias for brightness and contrast adjustment in the next step.

### 3.1.3. Brightness and contrast adjustment

In the last preprocessing step we use Otsu Binarization approach [7]. As a limitation of this algorithm it's assumes uniform illumination (implicitly). In our case this not true because we are dealing with text-photos produced with digital cameras. To bypass this issue, we suggest using a strategy of brightness and contrast adjustment.

Many approaches have been proposed for contrast enhancement and brightness control [9][10][11] [40]. But none of these can solve the problem we are handling. As a solution, we propose to use a simple yet efficient pixel transform to create an operator for brightness and contrast adjustment.

We multiply each input pixel with a parameter a>0 called gain and add a second parameter ß called bias to the resulting multiplication.

a is used to control the contrast

ß is used to control the brightness.

The equation of this operation is given by (16):

$$g(x, y) = \alpha \cdot f(x, y) + \beta \qquad\qquad (16)$$

f(x, y) is the source image.
g(x, y) is the resulting processed image.
Where (x, y) indicates that the pixel is located in the x-th row and y-th column.

Using the average brightness estimation calculated previously, we control the equation (3), so that the $\text{Brightness}_{\text{of}(g(x,y))} \leq 0.93$  ( $\alpha$ is an integer taking its values in $[0, 100]$).
We give the result of the proposed approach using:
$\alpha = 1.4 \; and \; \beta = 50$

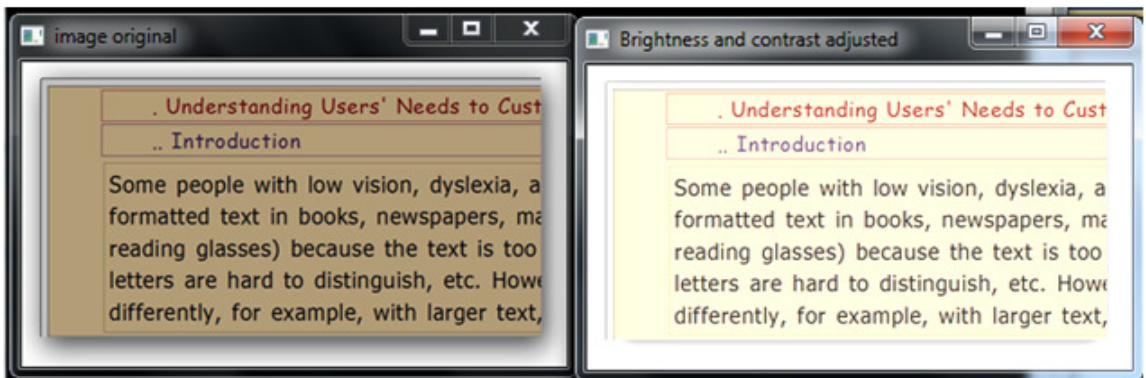

Figure 3: Left: the original image. Right: the resulting image using the proposed approach for brightness and contrast adjustment.

## 3.2. Grayscale conversion

Image involving only intensity are called intensity, gray scale, or gray level images. Grayscale conversion is one of the simplest image enhancement techniques. The main reason why grayscale representations are often used for extracting descriptors instead of operating on color images directly is that grayscale simplifies the algorithm and reduces computational requirements. Indeed, color may be of limited benefit in many applications and introducing unnecessary information could increase the amount of training data required to achieve good performance, that's the case for text recognition and identification. Many algorithms have been proposed for grayscale conversion. It' has been proven that not all color-to-grayscale algorithms work equally well [1], also it has been shown that Luminance algorithm perform better than other variations for texture based image processing [1]. In our case we use Luminance algorithm which is designed to match human brightness perception by using a weighted combination of the RGB channels in component-wise manner. Luminance is by far more important in distinguishing visual features [2]. Many algorithms exploit this property as for jpeg compression, where images are compressed in the YCbCr color space, and chrominance (Cb, Cr) are quantized and compressed more than luminance (Y) [3].

Luminance=0.3R+0.59G+0.11B, (17)

Where

R is the red value

G is the green value

And B is the blue value

Luminance does not try to match the logarithmic nature of human brightness perception, but this is achieved to an extent with subsequent gamma correction [4]. An example of the resulting image when applying this algorithm to convert images to grayscale level is given in (Figure 4)

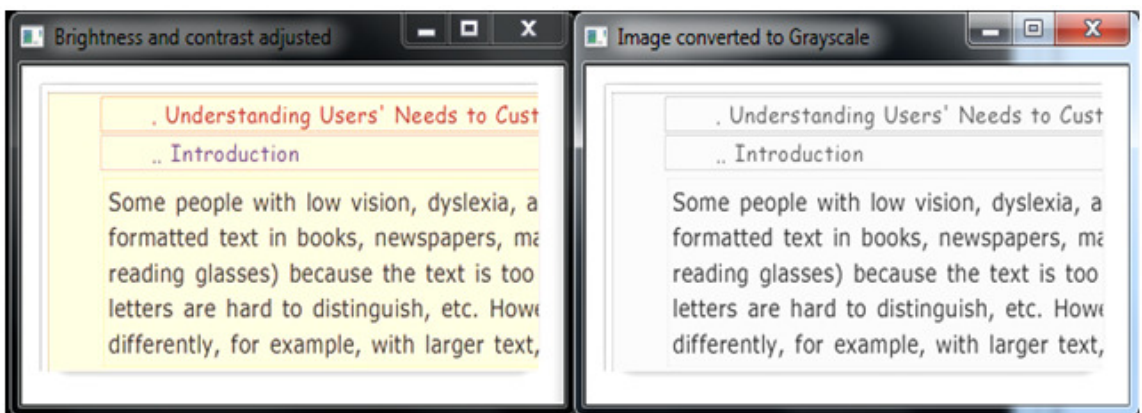

Figure 4: Left: the resulting image after brightness and contrast adjustment. Right: the image produced using the Luminance algorithm for grayscale conversion.

### 3.3. Un-sharp masking:

This step aims to enhance text details and edges by using Un-sharp masking filter. Sharpness describes the clarity of detail in a photo (document text in our case), and can be a valuable tool for emphasizing texture. Un-sharp masking filter also known as edge enhancement filter is a simple operator to enhance the appearance of detail by increasing small-scale acutance without creating additional detail [5][11]. The name was given because this operator improves details and other high frequency components in edge area via a process by subtracting a blurred version of the original image from the first one as illustrated in Figure 5.

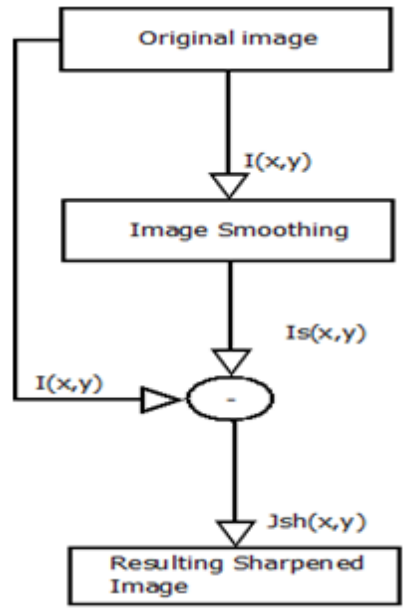

Figure 5: Block diagram of the classical Un-sharp masking

The principle of UM is quite simple [5] [6].

First a blurred version of the original image is created (we use a Gaussian blurring filter in our case). Then, this one is subtracted from the original image to detect the presence of edges, creating the unsharp mask. Finally this created mask is used to selectively increase the contrast of theses edges (fig. 4).

Mathematically this is represented by (18):

$$J_{sh}(x, y) = I(x, y) - I_s(x, y) \qquad (18)$$

Where $J_{sh}(x, y)$ is the sharpened resulting image
$I(x, y)$ is the original image
$I_s(x, y)$ is the smoothed version of $f(x, y)$ obtained by (19):

$$I_s(x, y) = I(x, y) - \{I(x, y) * HPF\} \qquad (19)$$

Where HPF is a height pass filter. Here we are using a Gaussian Kernel with 3x3 of size.

We give the result for Un-sharp masking using the parameters:
Amount=1.5;
Radius=0.5;
Threshold=0;

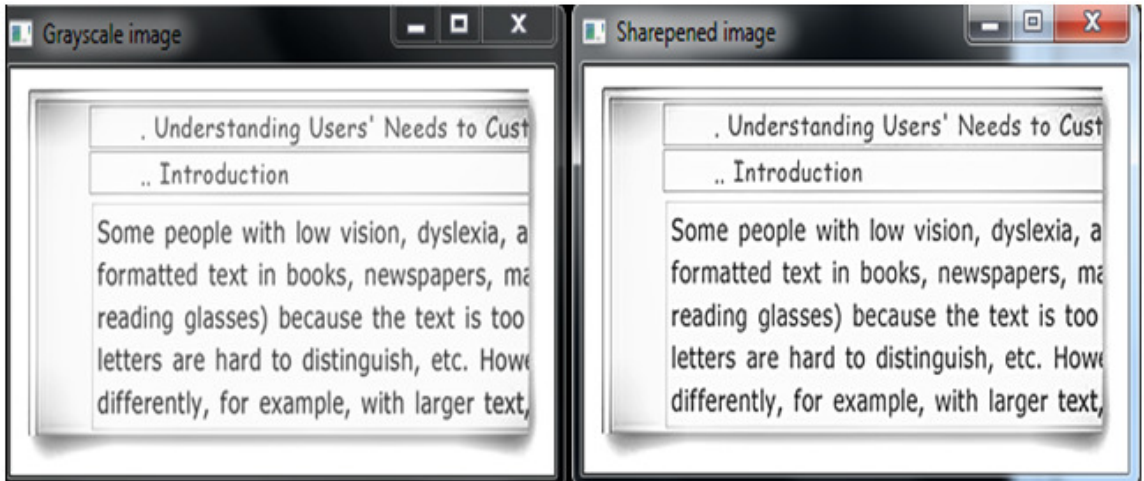

Figure 6: Left: the grayscale image produced in previous step. Right: the resulting sharpened image using Un-sharp Masking filter applied to the grayscale image in left.

As we can see in (Figure 6) Un-sharp masking is a very powerful method to sharpen images. But, too much sharpening can also introduce undesirable effects such as "halo artifacts". These are visible as light/dark outlines or halos near edges (Figure 7). Halos artifacts become a problem when the light and dark over and undershoots become so large that they are clearly visible at the intended viewing distance [11].

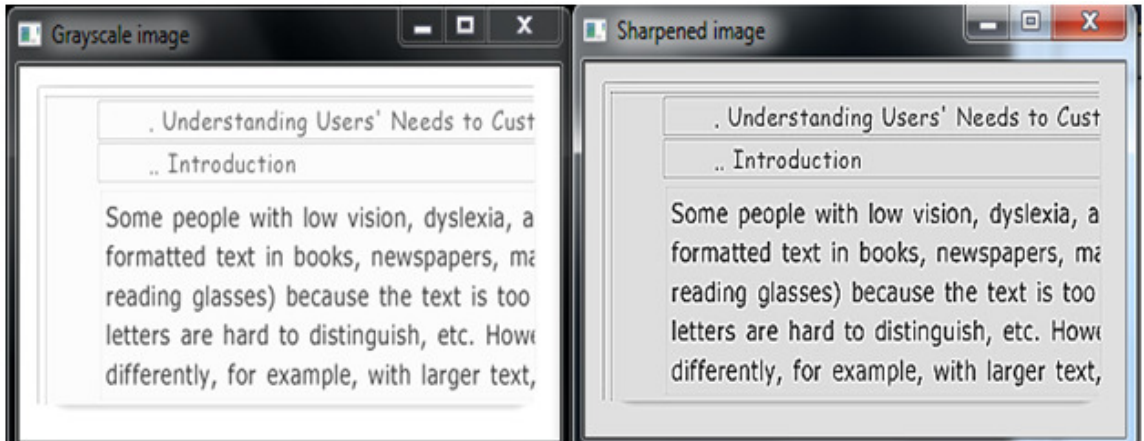

Figure 7: Left: the grayscale image produced in previous step. Right: the resulting sharpened image using Un-sharp Masking filter applied to the grayscale image in left with parameters (Amount=3,Radius=2.12,threshold=0).

### 3.4. Cleaning and whitening document background: Otsu thresholding

Thresholding is used to extract an object from its background by assigning an intensity value T (threshold) for each pixel such that each pixel is either classified as an object point or a background point.

Thresholding creates binary images from grey-level ones by turning all pixels below some threshold to zero and all pixels about that threshold to one. If g(x, y) is a threshold version of f(x, y) at some global threshold T, it can be defined as [8]

$$g(x, y) = \begin{cases} 1 \text{ if } f(x, y) \geq T \\ 0 \text{ otherwise} \end{cases} \tag{20}$$

Thresholding operation is defined as:

$$T = M[x, y, p(x, y), f(x, y)] \tag{21}$$

Where T is the threshold
f(x,y) is the gray value of point (x,y)
And p(x,y) is a local property of the point such as the average gray value of the neighborhood centered on point (x, y)
Converting a greyscale image to monochrome is an ordinary image processing task. Otsu's method [7] is an optimal thresholding, where a criterion function is devised that yields some measure of separation between regions. A criterion function is calculated for each intensity and that which maximizes this function is chosen as the threshold [7].
Otsu's thresholding chooses the threshold to minimize the intraclass variance (22) of the thresholded black and white pixels.

$$\sigma_w^2(t) = w_1(t)\sigma_1^2(t) + w_2(t)\sigma_2^2(t) \tag{22}$$

$w_i$ are the probabilities of the two classes separated by a threshold $t$ and $\sigma_i^2$ are variances of these classes.

It's based on a very simple idea: Find the threshold that minimizes the weighted within-class variance. This turns out to be the same as maximizing the between-class variance (23).

$$\sigma_b^2(t) = \sigma^2 - \sigma_w^2(t) = w_1(t)w_2(t)[\mu_1(t) - \mu_2(t)]^2 \tag{23}$$

This is expressed in terms of class probabilities $w_i$ and class means $\mu_i$
The class probability $w_1(t)$ is computed from the histogram as t:

$$w_1(t) = \sum_0^t p(i) \tag{24}$$

And the class mean $\mu_1(t)$ is:

$$\mu_1(t) = [\sum_0^t p(i) x(i)]/w_1 \tag{25}$$

Where x(i) is the value at the center of the ith histogram bin.

The goal is to find the threshold value where the sum of foreground and background spreads is at its minimum. (Figure 8) give the result of applying this algorithm of the previously preprocessed document.

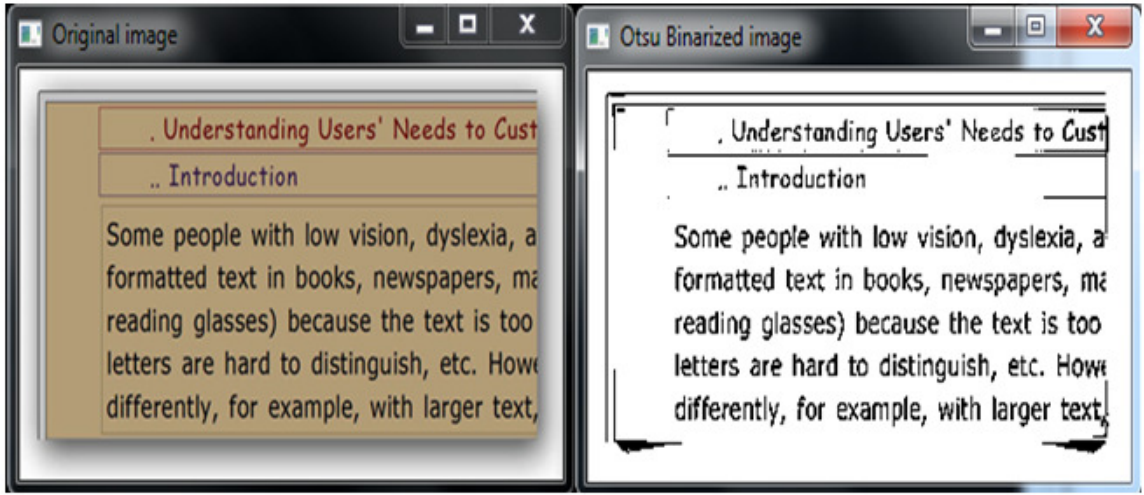

Figure 8: Left: the grayscale sharpened image produced in previous step using parameters (Amount=1.5, Radius=0.5, Threshold=0). Right: the resulting binarized image using Otsu thresholding approach.

Otsu algorithm assumes uniform illumination (implicitly), so the bimodal brightness behavior arises from object appearance differences only. This is not always true, especially for documents scanned using a digital camera. This can lead to a low OCR accuracy where the whole document can be ignored or show border effects (Figure 9). We can see that the produced binarized document image is very noisy. As we have predicted for this algorithm to behave efficiently in presence of non uniform illumination, we solved this problem using a linear strategy for brightness and contrast adjustment.

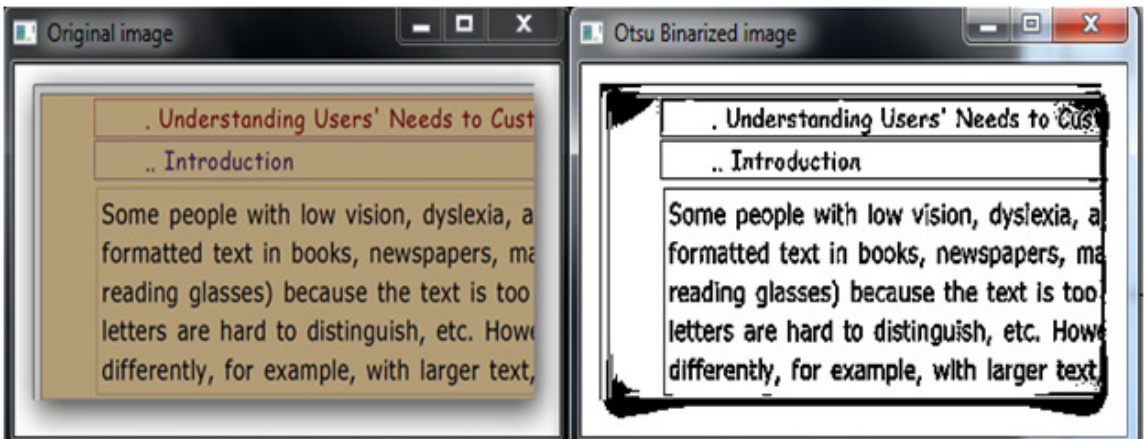

Figure 9: Left: the grayscale sharpened image produced in previous step using parameters (Amount=1.5, Radius=0.5, Threshold=0) without brightness and contrast adjustment. Right: the resulting binarized image using Otsu thresholding approach applied to the image in left.

# 4. PERFORMANCE EVALUATION

To demonstrate the efficiency of our proposed approach on improving the OCR accuracy, we test our method extensively against the standard database available on [42].

## 4.1. Test preparation

In this paper, the Open Computer Vision (OpenCV) is used as the implementation framework for Un-sharp masking and Otsu binrization, while we use our own implementation for Illumination adjustment, Brightness Equalization, Brightness estimation, Brightness and contrast adjustment, Grayscale conversion, HSV to RGB conversion and RGB to HSV conversion. Our approach is implemented using C++ language.

We test our approach on a PC with CPU: INTEL(R) PENTIUM(R) 2.13 GHZ dual core, RAM: 3GO and windows 7 Ultimate Edition (32 bits) as an operating system.

As a page-reader, we use the freely available software Tesseract-OCR (see table 1) bellow.

Table 1. Characteristics of the page-reader used in this paper.

| Developper | Version Name | Version Number | Platform | Version type | License |
|---|---|---|---|---|---|
| Ray Smith | tesseract-ocr | 3.02.0.2 | Windows 7 | Release | Freeware (Apache License, Version 2.0) |

We use Tesseract-OCR on the original images without applying our approach, then we apply our approach on the original images and we use Tesseract-OCR to measure the difference in performance.

Our test is based on the datasets available on [42], we used only the English version of the available document images. Table 2 gives the details of the tested document images.

Table 2. Test dataset details.

| Document ID | Document type | Page Number | Columns | Words | Characters |
|---|---|---|---|---|---|
| AR00233 | Magazine | 18 | 4 | 300 | 1516 |
| AR00235 | Magazine | 32 | 3 | 1201 | 6087 |
| AR00270 | Proceedings | 353 | 1 | 294 | 1367 |
| AR00319 | Journal | 99 | 1 | 701 | 3775 |

## 4.2. Accuracy Test results:

In this step the text generated by the used page-reading system is matched with the correct text to determine the minimum number of edit operations (character insertions, deletions, and substitutions) needed to correct the generated text. This gives us the number of $errors$. If there are $n$ characters in the correct text, then the character accuracy is defined by (26)

$$accuracy = \frac{n - errors}{n} \qquad (26)$$

Table. 3 shows the number of errors for our tests and the corresponding character Accuracy

Table 3: Character accuracy using Original and processed document image

| Document ID | Original image | | Processed image | |
|---|---|---|---|---|
| | Errors | Accuracy | Errors | Accuracy |
| AR00233 | 10 | 99.34% | 0 | 100% |
| AR00235 | 60 | 99.01% | 30 | 99.5% |
| AR00270 | 312 | 77.17% | 219 | 83.97% |
| AR00319 | 89 | 97.64% | 15 | 99.6% |

As the table shows, we have tested our approach on articles from, Magazine, Proceedings and journal.

As we can see in table 2, the number of columns in a document image or document type does not have a big effect on the final result, but the quality of the input image does.

We can conclude from the table that our method was very efficient on improving the OCR accuracy in all cases processed in this paper. The enhancement was from 2 percent to 6.8 percent.

We give the resulting output text for the processed document used to illustrate the approach presented in this paper.

(a) Original image

```
1 . Understanding Users' Needs to Cusf
7 .. Introduction

Some people with low vision, dyslexia,
aa
formatted text in books, newspapers,
ma
reading glasses) because the text is too
letters are hard to distinguish, etc. How:
differently, for example, with larger
text,
```

(b) Processed image

```
. Understanding Users' Needs to Cust
,. Introduction
Some people with low vision,
dyslexia, a
formatted text in books, newspapers,
me
reading glasses) because the text is
too
letters are hard to distinguish, etc.
Howt
differently, for example, with larger
text,
```

As we can see in this output text, the page reader was able to recognize all characters in the processed document image except for one character, whereas it was not able to recognize five characters in the original document image.

# 5. CONCLUSIONS

We have presented a novel approach for OCR Accuracy enhancing, the algorithm combine effectively a stack of very efficient preprocessing algorithm and adapt, illumination and brightness adjustment to our context of document image enhancement. The algorithm applies the Contrast Limited Adaptive Histogram Equalization separately on the Value of HSV color space to improve the Equalize the brightness of the original image. For Background cleaning, whitening and noise reducing we used Otsu Binarization algorithm as an aptimal Global thresholding algorithm.

Finally, we have demonstrated the efficiency of our approach by exhaustively testing it on a standards dataset [42], and presented some of the improvement introduced by the proposed approach on OCR Accuracy.

## ACKNOWLEDGEMENTS

This research was supported by the INVENTIVE Technologies laboratory[1] , a part of the Creargie MediaScan[2] in CASABLANCA Morocco and the RSAID Laboratory: "Remote sensing/Signal-image Processing & Applied mathematics/Informatics/ Decision making". The National School for Applied Sciences of Tetuan from university Abdelmalek Essaâdi. We are grateful to Mr. Dominique Schwartz the CEO of INVENTIVE Technologies and Creargie MediaScan for his valuable inputs, to Mme Sanaa Yassine for her valuable remarks and guidance, as well as to many other colleagues at RSAID Laboratory and INVENTIVE Technologies laboratory for their very helpful discussions.

---

[1]http://www.inventive-technologies.com/
[2]http://www.creargie.com/

## AUTHORS

El Harraj Abdeslam received his MS degree in Telecommunications and Network engineering from the National School of Applied Sciences, Tangier, Morocco in 2008 and an MS degree in Entreprise management from University of Perpignan, French in 2010. Since 2012 he is a PhD. Student in computer vision with University Abdelmalek Essaâdi, Tangier-Tetuan, Morocco. From 2008 to 2012, he was a Programmer Engineer with Creargie Maroc Company, Casablanca, Morocco. Currently, he is heading the Research and Development Department in the Company INVENTIVE Technologies, Casablanca, Morocco and also a member of Remote-Sensing & Mobile-GIS Unit/Telecoms Innovation & Engineering Research group.

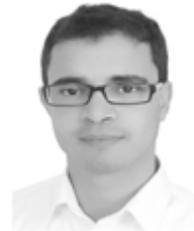

Raissouni Naoufal received the M.S., and Ph.D. degrees in physics from the University of Valencia, Spain, in 1997, and 1999, respectively. He has been a Professor of physics and remote sensing at the National Engineering School for Applied Sciences of the University Abdelmalek Essaadi (UAE) of Tetuan, since 2003. He is also heading the Innovation & Telecoms Engineering research group at the UAE, responsible of the Remote Sensing & Mobile GIS unit. His research interests include atmospheric correction in visible and infrared domains, the retrieval of emissivity and surface temperature from satellite image, huge remote sensing computations, Mobile GIS, Adhoc networks and the development of remote sensing methods for land cover dynamic monitoring.

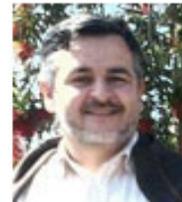